\documentclass[11pt,a4paper]{article}
\usepackage[hyperref]{acl2020}
\usepackage{times}
\usepackage{latexsym}
\usepackage{graphicx}
\usepackage{multirow}
\usepackage{booktabs}
\usepackage{comment}
\usepackage{caption}
\usepackage{subcaption}
\usepackage{natbib}
\usepackage{amsmath, bm, mathtools}
\usepackage{CJKutf8}
\graphicspath{ {./sections/figs/} } 

\usepackage[symbol]{footmisc}

\usepackage{microtype}

\aclfinalcopy 

\newcommand{\veryshortarrow}[1][3pt]{\mathrel{%
   \hbox{\rule[\dimexpr\fontdimen22\textfont2-.2pt\relax]{#1}{.4pt}}%
   \mkern-4mu\hbox{\usefont{U}{lasy}{m}{n}\symbol{41}}}}

\title{A Study of Cross-Lingual Ability and Language-specific Information in Multilingual BERT}

\author{Chi-Liang Liu\footnotemark[1] \quad Tsung-Yuan Hsu\footnotemark[1] \quad Yung-Sung Chuang\footnotemark[1] \quad Hung-Yi Lee\\
College of Electrical Engineering and Computer Science, National Taiwan University \\
{ \tt \{sivia89024, liangtaiwan1230, tlkagkb93901106\}@gmail.com} \\
{ \tt b05901033@ntu.edu.tw}\\}

\date{}

\begin{document}
\maketitle
\footnotetext[1]{Equal Contirbution}
\footnotetext[2]{This work was supported by Delta Electronics, Inc..}
\begin{abstract}
Recently, multilingual BERT works remarkably well on cross-lingual transfer tasks, superior to static non-contextualized word embeddings. In this work, we provide an in-depth experimental study to supplement the existing literature of cross-lingual ability. We compare the cross-lingual ability of non-contextualized and contextualized representation model with the same data. We found that datasize and context window size are crucial factors to the transferability. We also observe the language-specific information in multilingual BERT. By manipulating the latent representations, we can control the output languages of multilingual BERT, and achieve {\it unsupervised token translation}. We further show that based on the observation, there is a computationally cheap but effective approach to improve the cross-lingual ability of multilingual BERT.  
\end{abstract}
\section{Introduction}
Recently, multilingual BERT (m-BERT) \cite{devlin:19} has shown its superior ability in cross-lingual transfer on many downstream tasks, either in a way it is used as a feature extractor or finetuned end-to-end~\citep{conneau:18, wu:19, hsu-etal:19, pires:19}. 
It seems that m-BERT has successfully learned a set of cross-lingual representations in a shared vector space for multiple languages. However, given the way how m-BERT was pre-trained, it is unclear how it succeeded in building up cross-lingual ability without parallel resources and learning on supervised objectives explicitly.

Some key components of BERT model architecture, such as {\it depth} and {\it total number of parameters}, have been found correlated with the cross-lingual ability of m-BERT~\citep{Cao:20}, but there are still some critical factors in pretraining yet to be recognized. For example, ~{\it massive amount of data} and {\it relatively long context window when training} are two key components specific to m-BERT but haven't been discussed in the literature.   
In Section~\ref{section:align}, we study the impacts of these components on the cross-lingual ability of m-BERT to enrich our understandings of how to build a powerful cross-lingual model.  

Also, there is another question we want to answer in this work. It has been widely proven that m-BERT's cross-lingual ability is related to its success in aligning cross-lingual word pairs, but does language-specific information still exist in the representation of m-BERT? 
In Section~\ref{section:trans}, we show that language-specific information is still encoded in m-BERT.
By just adding one fixed vector to all latent representations, we can make m-BERT output sentences in another language semantically close to English input sentences.
Moreover, this allows us to boost the zero-shot transferability of m-BERT without any extra efforts.

The contributions of this work can be summarized as the following:
\begin{itemize}
    \item We evaluated the cross-lingual ability of m-BERT influenced by different datasizes and window sizes.
    \item We compared the cross-lingual ability of m-BERT with non-contextualized word embedding trained. 
  \item We developed an approach to extract language-specific representations in m-BERT, so we can make m-BERT decode a different language from input language.
  \item We improved the performance of m-BERT on a cross-lingual transfer downstream task by language-specific representations.
\end{itemize}
\section{Related Work}
\textbf{Multilingual BERT.} BERT~\citep{devlin:19} is a Transformer-based~\citep{vaswani:17} large pre-trained language model that has been widely applied in numerous NLP tasks, showing great potentials in transfer learning. And m-BERT, which is pre-trained on Wikipedia text from 104 languages without cross-lingual objective or parallel data, serves as a good cross-lingual representation model that generalizes well across languages for a variety of downstream tasks~\cite{wu:19, hsu-etal:19, pires:19}. There are a line of works studying the key components contributing to the cross-lingual ability of m-BERT~\citep{Karth:20, tran:20, Cao:20, singh:19}, sometimes coming with inconsistent observations, showing that our understandings about it are still in the early stages.\\

\noindent\textbf{Cross-lingual Word Embedding.} The goal of cross-lingual word embedding is to learn embeddings in a shared vector space for two or more languages. A line of works assumes that monolingual word embeddings share similar structures across different languages and try to impose post-hoc alignment through a mapping~\citep{mikolov:13, Smith:17, joulin:18, Lample:18, artetxe:18, Zhou:19}. Another line of works considers joint training, which optimizes monolingual objective with or without cross-lingual constraints when training word embeddings~\citep{luong:15,gouws:15,Ammar:16,duong:16,lample:18b}. Cross-lingual word embedding methods above were initially proposed for non-contextualized embedding such as GloVe~\citep{pennington:14} and Word2Vec~\citep{mikolov:13w2v}, and later adapted to contextualized word representation~\citep{schuster:19, aldarmaki-diab:19}. Recently, token representations from m-BERT also serve a similar role as cross-lingual word embedding in many works.

\section{How to Build up Cross-lingual Ability}
\label{section:align}
Cross-lingual alignment of words in m-BERT representation has been observed and credited with cross-lingual transferability~\citep{Cao:20}. The transferbility come from semantically similar words are encoded into similar latent representations, regardless of languages. Considering  that m-BERT was trained on masked language modeling and there are a bunch of subwords shared by languages that could be used as anchors, it is hypothesized that m-BERT exploited the co-occurrence information to align word pairs having similar contexts or usage statistics~\citep{pires2019multilingual}. 

However, m-BERT seems to be the only model succeeding in aligning cross-lingual word pairs simply trained on concatenated monolingual corpora with a monolingual objective. Similar methods for training cross-lingual word embeddings, called unsupervised joint training, often fail to generate high-quality alignments, especially for embeddings of non-shared words, due to a lack of explicit clues~\citep{Wang:20}. \citet{Wang:20} show that supervised post-hoc alignment effectively boosts unsupervised joint training in the case of traditional non-contextualized embeddings, but is not as beneficial in the case of m-BERT, suggesting that pretraining alone generates strong alignments even without supervision. 

Several key components have been studied why m-BERT generalize across languages so well~\citep{Karth:20}. In an analysis of the two-language version of BERT, it was shown that depth and the total number of network parameters affect cross-lingual ability a lot. Performance on the XNLI dataset significantly degrades when model architecture becomes shallower or more-parsimonious. Wordpiece tokenization also contributes to cross-lingual ability, compared with character-level or word-level tokenization. 

Surprisingly, shared vocabulary was found to be unnecessary in the same study. Pre-trained BERT successfully transfers from English to a lexical-modified version of English, although it failed when word ordering or n-gram statistics were randomly manipulated. This suggests that some structural similarity is possibly what BERT has learned from, while further definition should be developed and discovered. 

It's also unclear if the above findings hold for other non-contextualized embedding models, since the pre-training process and dataset could make a crucial difference in cross-lingual ability. 

In this section, we calibrate the external factors when pretraining the representations to give a fair discussion focusing on the architecture of the embedding models and study two key components that are dominant in building up cross-lingual ability of m-BERT.  

\subsection{Metrics for Cross-lingual Ability}
There are two main paradigms for evaluating cross-lingual representations:~{\it word retrieval} and {\it downstream task transfer}. 
Here we use both word retrieval and downstream task transfer as the indicators of cross-lingual ability for both contextualized and non-contextualized embeddings. 
Although word retrieval is a task originally proposed to measure cross-lingual alignment at the word level, therefore naturally more suitable for non-contextualized embeddings, contextual version of word retrieval has been proposed for contextualized embeddings and consistent with downstream task transfer performance~\citep{Cao:20}. 

\subsubsection{Word Retrieval}
Given a word and a bilingual dictionary~$D=\{(x_1,y_1),(x_2,y_2),\ldots,(x_n,y_n)\}$, listing all parallel word pairs from source and target languages, word retrieval is the task to retrieve the corresponding word in target language considering information provided by embedding vectors~$\{(u_1,v_1),(u_2,v_2),...(u_n,v_n)\}$. Specifically we consider a nearest neighbor retrieval function
\begin{equation}
    neighbor(i)=\mathop{\arg\max}_{j}sim(u_i,v_j),
\end{equation}
where $u_i$ is the embedding of source word $x_i$ and we want to find its counterpart $y_i$ among all candidates~$y_1,y_2,...y_n$. And we use cosine similarity as the similarity function $sim$ .

Then we have{\it~mean reciprocal rank }(MRR) as evaluation metrics. MRR is defined as
\begin{equation}
mean~reciprocal~rank = \frac{1}{n}\sum_{i}^{n}\frac{1}{rank(y_i)}
\end{equation}
where $rank(y_i)$ is a ranking function based on retrieval results.
For contextualized embeddings, we simply average embeddings in all contexts and use the mean vector to represent each word, so that contextualized embeddings could also be evaluated with the task defined above.

\subsubsection{Downstream Task Transfer}
We consider XNLI as our downstream task to evaluate cross-lingual transfer. The XNLI dataset was constructed from the English MultiNLI dataset by keeping the original training set but human-translating development and test sets into other 14 languages~\citep{conneau:18, williams:18}. Given a pair of sentences, the task is to predict the relation of the sentence pair among three classes: entailment, neutral, or contradiction. As there are only training data in English, models should perform zero-shot cross-lingual transfer on development and test sets.

\subsection{Experiments}
To compare non-contextualized and contextualized embedding models, we conducted experiments with GloVe, Word2Vec and, BERT, where the number of dimensions was all set to 768. Embeddings were first pretrained from scratch and then evaluated on word retrieval and XNLI, to assess their cross-lingual ability.   

For pretraining data, we used Wikipedia from 15 languages~(English, French, Spanish, German, Greek, Bulgarian, Russian, Turkish, Arabic, Vietnamese, Thai, Chinese, Hindi, Swahili, and Urdu) to pre-train all word embeddings following unsupervised joint training scenario, assuring each target language in the downstream task has been well pre-trained. 

For word retrieval task, we evaluated cross-lingual alignment between English and each of the remaining 14 languages, using bilingual dictionaries from MUSE\footnote{https://github.com/facebookresearch/MUSE}.
For XNLI zero-shot transfer task, the training set is in English, and the target languages of testing sets are the same as those used in the word retrieval task.  

To eliminate the effect of tokenization, we tokenized data uniformly with the vocabulary of m-BERT, especially the cased one, which is essentially wordpiece tokenization. 

Also, we experimented with different amounts of data: {\it 200k} and {\it 1000k} sentences per language to study the results under different data sizes. 
The results with {\it 200k} and {\it 1000k} sentences are shown in Sections~\ref{subsubsection:small} and~\ref{subsubsection:big}, respectively.  

\subsubsection{Small Pretraining Data}
\label{subsubsection:small}
\begin{figure}[ht!]
\centering 
\includegraphics[width=\linewidth]{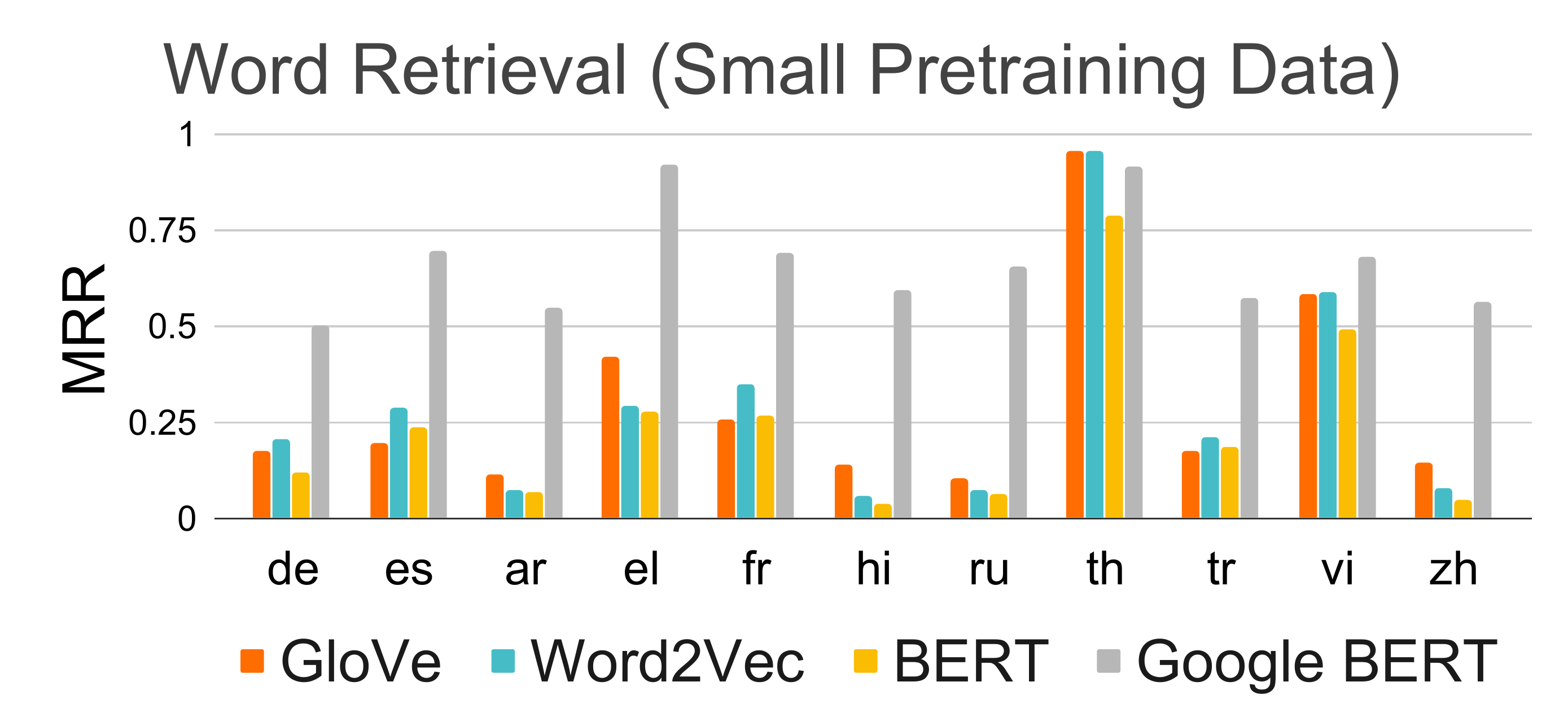}
 \caption{Evaluating alignment with Word Retrieval, compared between different representation models. For reference, Google BERT is the m-BERT pre-trained by Google. Except for Google Bert, other embeddings were pre-trained on the same data (200k sentences per language).}
 \label{fig:3_1}
 \vspace{-0.25cm}
\end{figure}
\begin{figure}[ht!]
\centering 
\includegraphics[width=\linewidth]{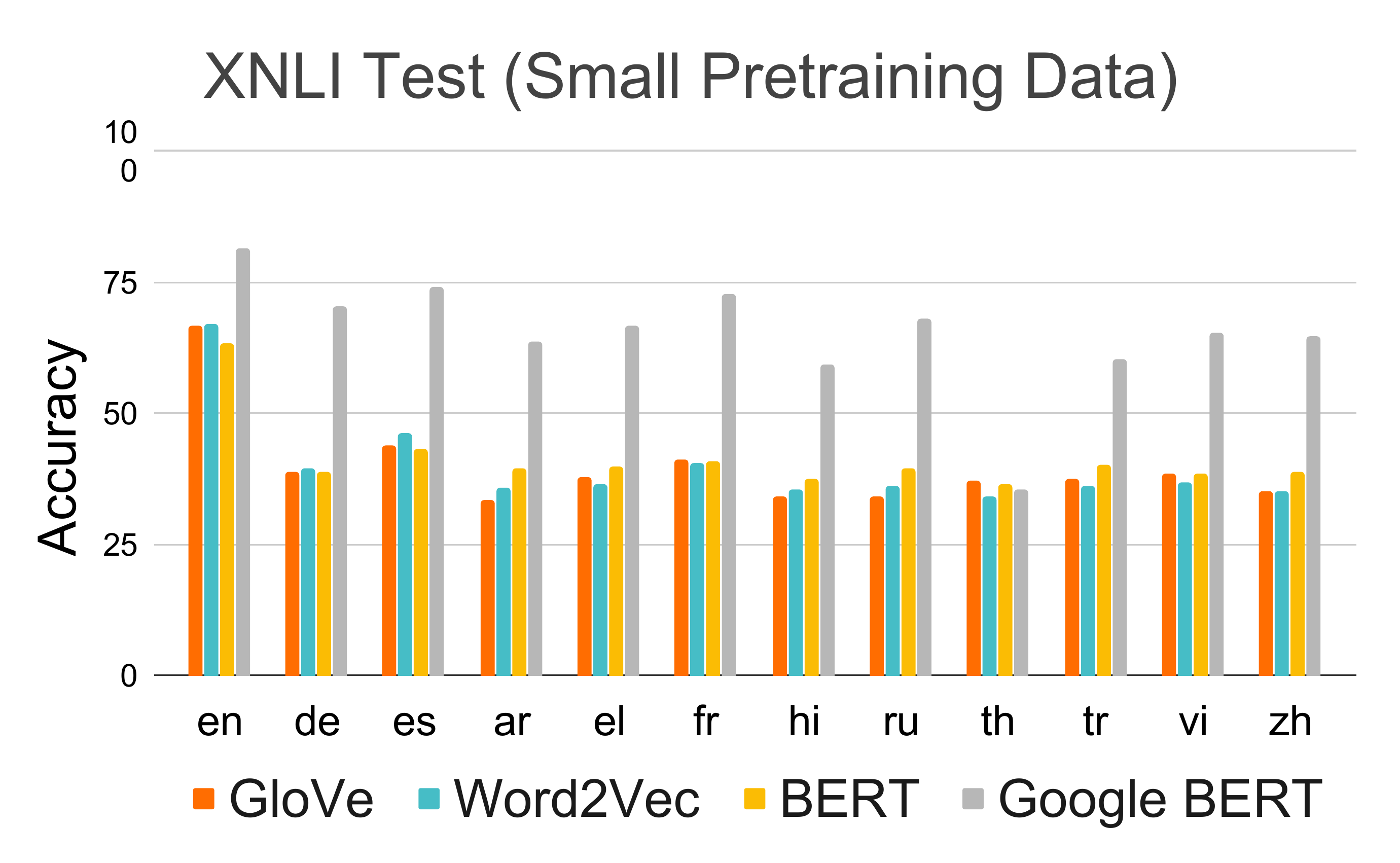}
 \caption{Performance comparison on XNLI zero-shot cross-lingual transfer task}
 \label{fig:3_2}
\end{figure}

Surprisingly, when pre-trained on small pretraining data (200k sentences per language), BERT didn't show its extraordinary cross-lingual ability as shown in Figure~\ref{fig:3_1} and \ref{fig:3_2}. For example, GloVe and Word2Vec achieved stronger cross-lingual alignment than BERT in terms of MRR score on word retrieval task on every language paired with English. And although BERT achieved better accuracies on XNLI zero-shot transfer on several languages, the margins were very small, and the overall performances were not better than GloVe and Word2Vec.  

This finding provides a further discussion of the literature. Work by ~\citet{Karth:20} has found that depth and the total number of network parameters were two decisive architecture elements dominating the cross-lingual ability of BERT, also supported by experiments on XNLI. 
When the number of attention heads and the number of total parameters were fixed, decreasing model depth degenerated cross-lingual transfer performance. On the other hand, when the number of attention heads and depth were fixed, decreasing the number of network parameters degenerated performance, either. 
It seems that the expressiveness or capacity of models is crucial to building up cross-lingual ability. 

However, is it really the case that the bigger, the better? 
From results in this experiment, the capacity of BERT is definitely much bigger than GloVe and Word2Vec, but in the case of pretraining on the limited size of data, BERT didn't achieve superior performance as expected, suggesting that the relation of model capacity and cross-lingual ability may not be monotonic and the size of pretraining data also comes into play. 

\vspace{-0.25cm}
\subsubsection{Big Pretraining Data}
\label{subsubsection:big}
\begin{figure}[ht!]
\centering 
\includegraphics[width=\linewidth]{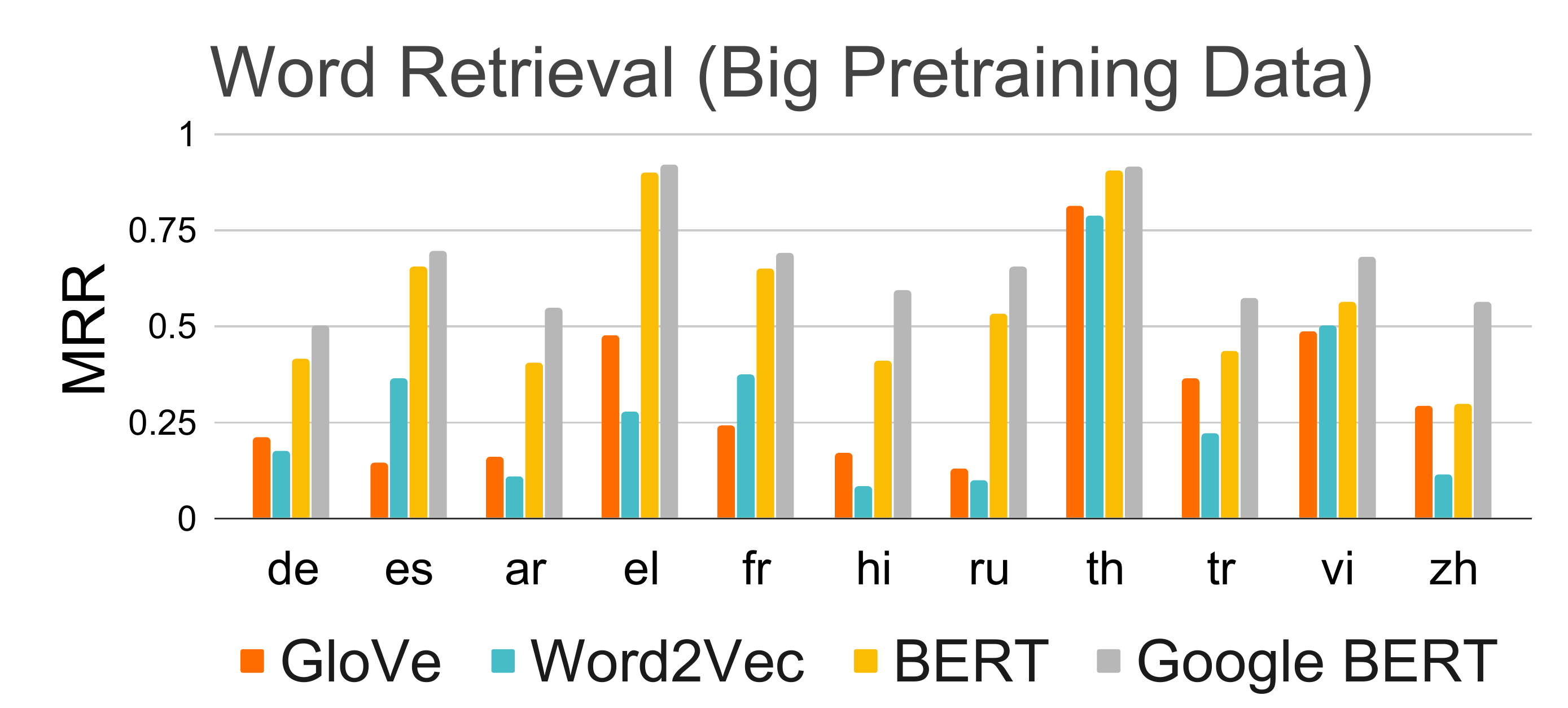}
 \caption{Evaluating alignment with Word Retrieval, compared between different representation models. For reference, Google BERT is the m-BERT pre-trained by Google. Except for Google BERT, other embeddings were pre-trained on the same data (1000k sentences per language).}
 \label{fig:3_3}
 \vspace{-0.25cm}
\end{figure}
\begin{figure}[ht!]
\centering 
\includegraphics[width=\linewidth]{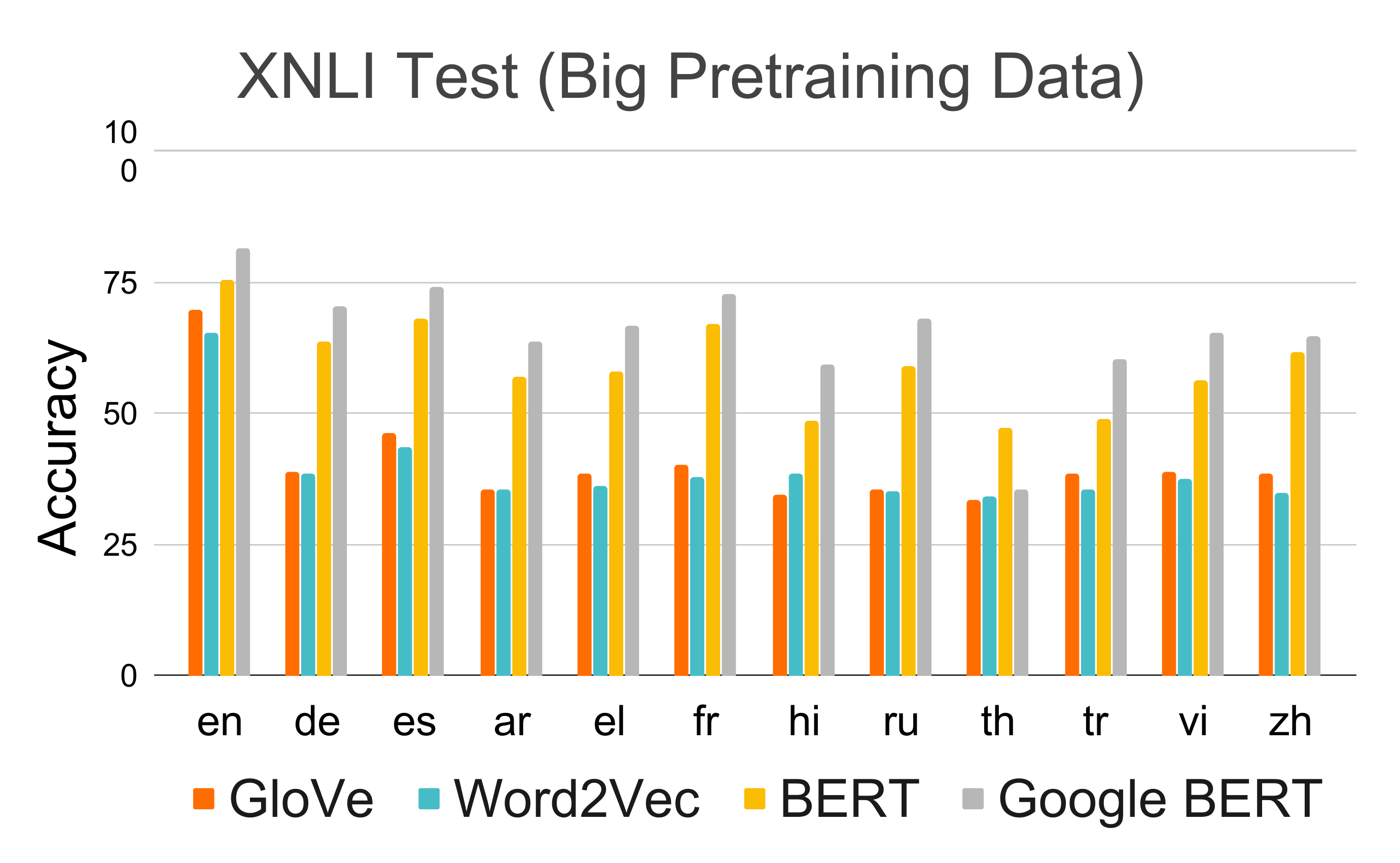}
 \caption{Performance comparison on XNLI zero-shot cross-lingual transfer task}
 \label{fig:3_4}
 \vspace{-0.25cm}
\end{figure}

When pre-trained on big pretraining data (1000k sentences per language), there was a dramatic turn as shown in Figure~\ref{fig:3_3} and~\ref{fig:3_4}. BERT achieved an overwhelmingly higher MRR score than other embeddings on every XX-En language pairs, showing that it did a much better job in aligning semantically similar words from different languages. 

Testing results on XNLI were also consistent with word retrieval task, BERT reached higher accuracies than GloVe and Word2Vec, demonstrating that it had the better cross-lingual ability.  

It was noticeable that the increase in pretraining data size largely improved the cross-lingual alignment and transferability of BERT, while it was not the same case for GloVe and BERT. And the bounding performance of Google BERT, which is the pretrained parameters released by Google, shows that there is still room for improvement if given even more pretraining data.

Besides the effect of amount of data, in the next subsection, we further studied other key components making BERT capable of learning a good cross-lingual representation space.

\subsubsection{Breaking Down Long Dependency}
We noticed that the typical co-occurrence window size of non-contextualized embeddings, like GloVe and Word2Vec,  are often limited to 5$\sim$30 tokens, but BERT could attend to hundreds of tokens, which means that BERT could learn from longer dependency and richer co-occurrence statistics. So we experimented with a smaller window size to find out if longer dependency is also necessary for learning cross-lingual structures.    
We directly sliced sentences in original pretraining data into smaller segments, limiting input length to 20 tokens for each example. And then we evaluated embeddings pretrained on these segments for cross-lingual ability\footnote{Limiting the number of tokens attended by attention heads may not work because the information from far tokens could still flow through layers and be collected at deeper layers.}.

\begin{figure}[ht!]
\begin{subfigure}{\linewidth}
  \includegraphics[clip,width=\linewidth]{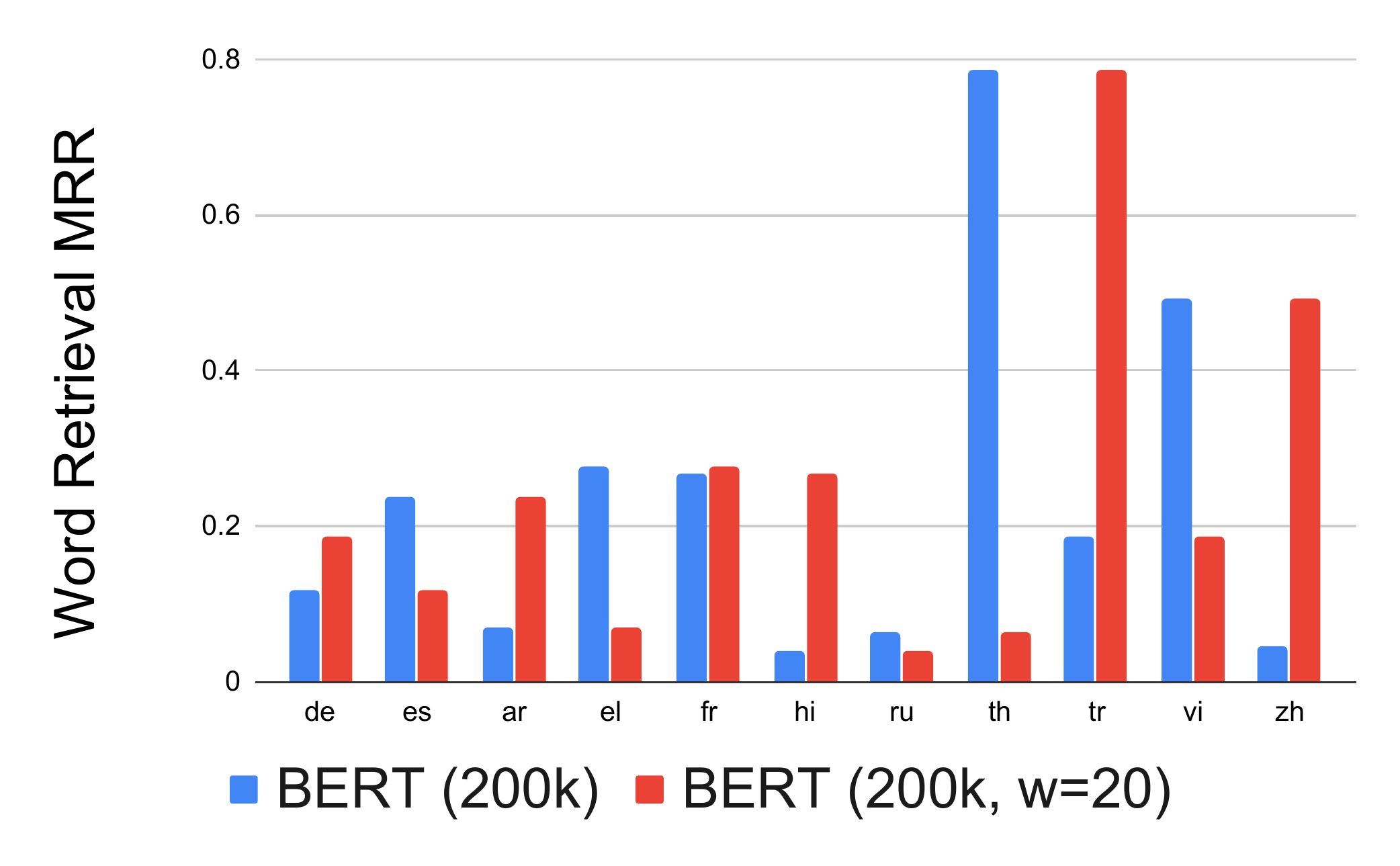}%
\end{subfigure}
\begin{subfigure}{\linewidth}
  \includegraphics[clip,width=\linewidth]{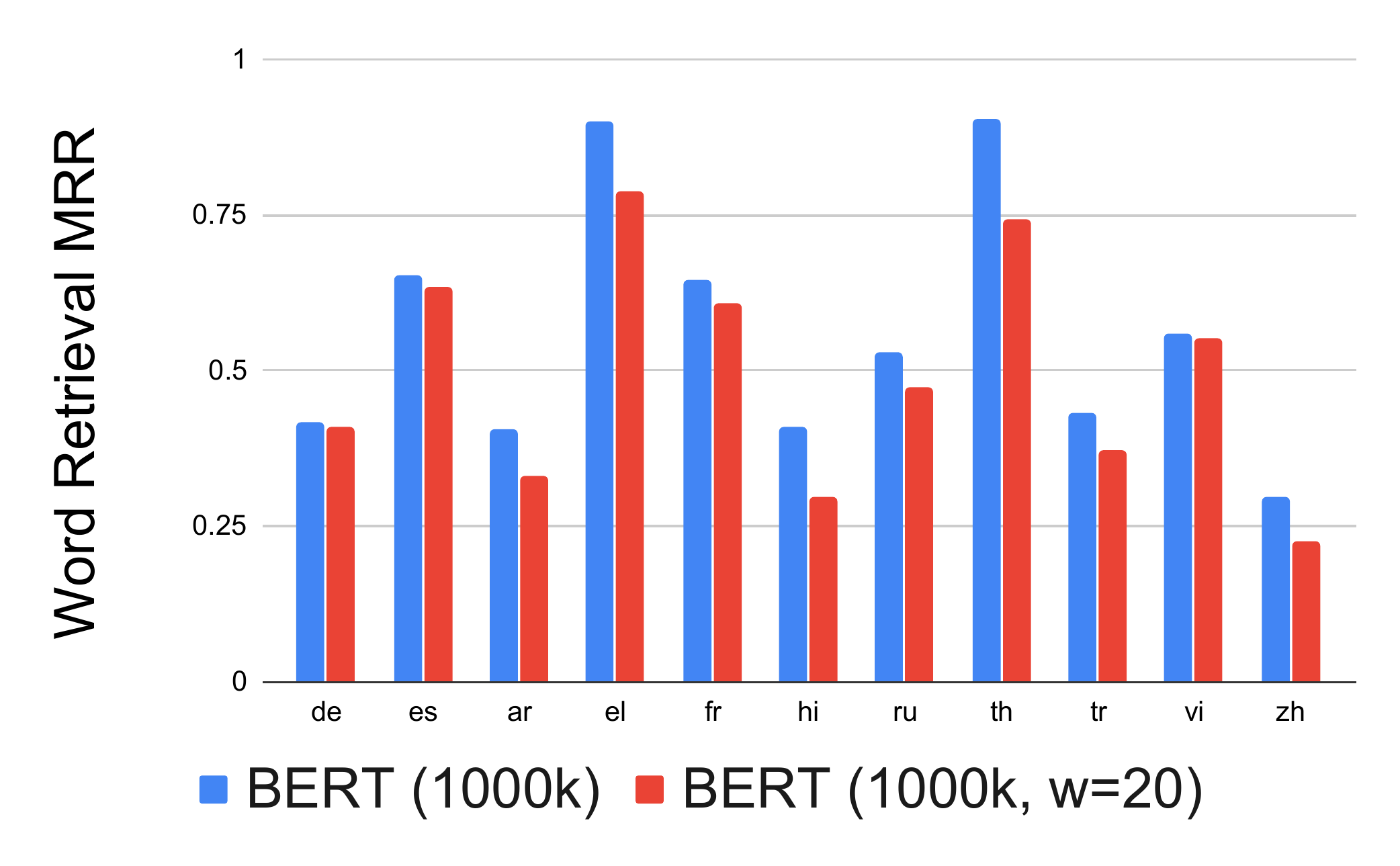}%
\end{subfigure}
\caption{The Effect of Window Size}
\label{fig:3_5}
\vspace{-0.25cm}
\end{figure}

Results are shown as figure~\ref{fig:3_5}. In the case of big pretraining data, pretraining BERT with shortened inputs drastically hurt the cross-lingual ability of BERT, indicated by lower MRR score on word retrieval task compared to BERT pre-trained on normal-lengthed data. It should be noticed that the total number of tokens in pretraining data stayed unchanged.

However, in the case of small pretraining data, pretraining BERT with shortened inputs yielded to better cross-lingual alignment on several languages, suggesting that breaking down long dependency helps BERT to learn better cross-lingual alignment when only limited data is available.  

\begin{figure}[ht]

\begin{subfigure}[b]{\linewidth}
  \includegraphics[clip,width=\linewidth]{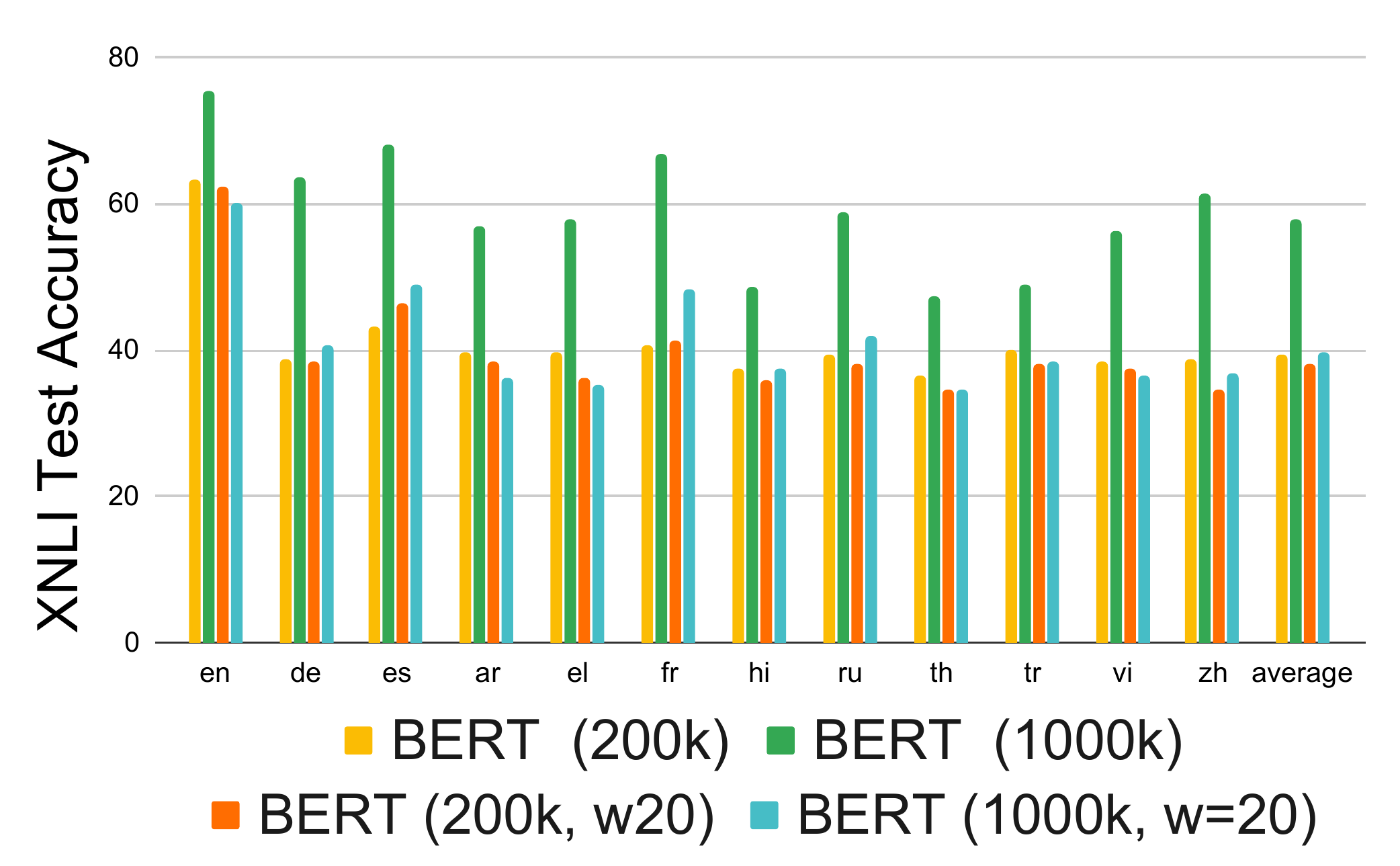}
  \subcaption{When limiting context window, the effect of pretraining datasize is less significant on XNLI transfer performance}
\end{subfigure}

\begin{subfigure}[b]{\linewidth}
  \includegraphics[clip,width=\linewidth]{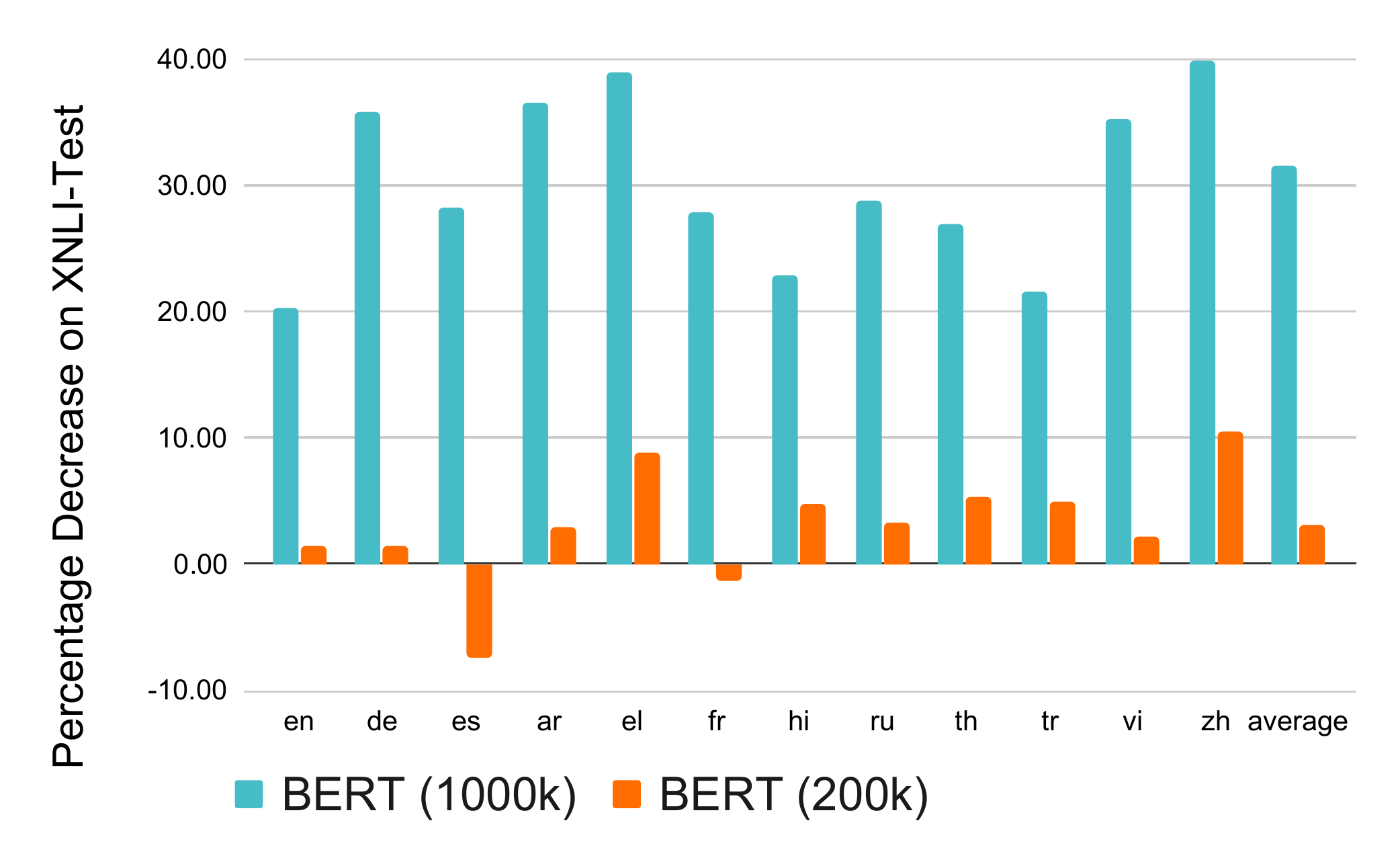}
  \subcaption{Performance drop on XNLI due to limiting context window size (original 512$\rightarrow$20) is shown in percentage. It is shown that the performance decrease of BERT pre-trained on small data is less than big data and even negative on some languages, which means there is an increase in transfer}
\end{subfigure}
\caption{The Effect of Window Size}
\label{fig:3_6}
\vspace{-0.25cm}
\end{figure}

In Figure~\ref{fig:3_6},  it is shown that when there were no long dependencies to learn, the benefit of increasing the datasize became minor. 
The results on XNLI also show that breaking down long dependency was more harmful when there were more pretraining data.   

Considering the above observations, we hypothesized that cross-lingual alignment or cross-lingual transferability of BERT are learned not only from local co-occurrence relations but also from co-occurrence relations of global scope, with a larger amount of data and model capacity required.

\section{Language Information}
\label{section:trans}
In this section, we first analyzed m-BERT using the mean difference between 15 languages, and then show that m-BERT can decode other languages by modifying the hidden representation.
\subsection{Language Information in BERT}
Although, in Section~\ref{section:align} we showed that m-BERT does extremely well on aligning cross-lingual representation, 
the representations of tokens from different languages are still able to be distinguished easily. For example, when masking one token from an English sentence, the probability of decoding a Chinese token in that position is very low for m-BERT model. This phenomenon is largely different from non-contextualized word representation alignment, which makes different language indistinguishable in the embedding space. We believed that m-BERT reserves some implicit language-specific information in the embedding space that can be disentangled from semantic embedding. As a result, we wanted to find out the language-specific component in the hidden representation.
%

\begin{table*}[t]
\centering
\caption{Unsupervised Token Translation random sample (applied MDS on layer 10)}
\label{tab:ex}
\footnotesize
\setlength\tabcolsep{3pt}
\begin{CJK*}{UTF8}{gbsn}
\begin{tabular}{r|l}
\toprule
Input (en) & {The girl that can help me is all the way across town. There is no one who can help me.}\\
\midrule
Ground Truth (zh) & 能帮助我的女孩在小镇的另一边。 没有人能帮助我。。 \\
en$\veryshortarrow$zh, $\alpha=1$ & . 孩 ， can 来 我 是 all the way across 市 。 。 There 是 无 人 人 can help 我 。\\
en$\veryshortarrow$zh, $\alpha=2$ & . 孩 的 的 家 我 是 这 个 人 的 市 。 。 他 是 他 人 人 的 到 我 。\\
en$\veryshortarrow$zh, $\alpha=3$ & 。 ， 的 的 的 他 是 的 个 的 的 ， 。 ： 他 是 他 人 ， 的 。 他 。\\
\midrule
Ground Truth (fr) & La fille qui peut m'aider est à l'autre bout de la ville. Il n'y a personne qui pourrait m'aider.\\
en$\veryshortarrow$fr, $\alpha=1$ & . girl qui can help me est all la way across town . . There est no one qui can help me .\\
en$\veryshortarrow$fr, $\alpha=2$ & . girl qui de help me est all la way dans , . . Il est de seul qui pour aid me .\\
en$\veryshortarrow$fr, $\alpha=3$ & , , , de , me , all la , , , , , n n n n , , , , ,\\
\bottomrule
\end{tabular}
\end{CJK*}
\end{table*}

\subsubsection{Language-specific Representation}
We assumed that we have $n$ languages denoted by $\{L_1, L_2, \ldots, L_n\}$ and their corresponding corpora. For each language, there is a language-specific representation $R^l_{L_i}$ in each layer $l$. 

To describe our method of finding language-specific representations and the use of language-specific representations, we defined the following notation
 \begin{itemize}
     \item \textbf{Context-Dependent Representation:} Given an input sequence $x$ and token index $i$, we denote the hidden representation in layer $l$ by $\bm{h}^l_{x,i}$.

     \item \textbf{Mean of Language:} Given a language $L$ and its corresponding corpora $C$ contain $n$ inputs $x_1, x_2,\ldots x_n$, we denote the language-specific representation of layer $l$ by $$R^l_{L}=\mathop{E}_{x \in C, \\ i \in I}\left[\bm{h}^l_{x,i}\right]$$ which represents the mean of all the context-dependent representation in the corpora.
     \item \textbf{Mean Difference Shift:} Given two languages $L_1, L_2$, we defined the mean difference shift of layer $l$ as $$D^l_{L_1\veryshortarrow L_2}=R^l_{L_2}-R^l_{L_1}$$
 \end{itemize}

In this paper, as the first exploration on language-specific representation, we simply took Mean of Language (MoL) as the language-specific representation and used Mean Difference Shift (MDS) to swift the representation from one language to another language.

\begin{table*}[t]
\centering
\caption{Best BLEU-1 score of mean difference shift on Unsupervised Token Translation with best $\alpha$ is selected from $[1.0, 2.0, 3.0]$, and the best layer is selected from $[1, 12]$.}
\label{tab:mt}
\footnotesize
\setlength\tabcolsep{3pt}
\begin{tabular}{c|cccccccccccc}
\toprule
& en$\veryshortarrow$de & en$\veryshortarrow$fr & en$\veryshortarrow$ur & en$\veryshortarrow$sw & en$\veryshortarrow$zh & en$\veryshortarrow$el & de$\veryshortarrow$en & fr$\veryshortarrow$en & ur$\veryshortarrow$en & sw$\veryshortarrow$en & zh$\veryshortarrow$en & el$\veryshortarrow$en \\
\midrule
BLEU-1 (layer=11, $\alpha$=3) & 8.01 & 12.18 & 6.32 & 9.11 & 19.68 & 11.47 & 9.24 & 8.30 & 9.35 & 7.92 & 8.58 & 5.24 \\
\midrule
best BLEU-1 & 12.35 & 12.18 & 7.83 & 9.11 & 23.21 & 19.13 & 10.44 & 12.27 & 9.35 & 9.86 & 8.58 & 9.07 \\
\midrule
best layer & 10 & 11 & 7 & 11 & 7 & 10 & 1 & 10 & 11 & 9 & 11 & 7 \\
best $\alpha$ & 3.0 & 3.0 & 2.0 & 3.0 & 2.0 & 3.0 & 3.0 & 3.0 & 3.0 & 3.0 & 3.0 & 3.0 \\
\midrule
convert rate, (layer=10, $\alpha$=1) & 0.402 & 0.417 & 0.611 & 0.153 & 0.478 & 0.621 & 0.452 & 0.496 & 0.299 & 0.147 & 0.239 & 0.302\\
\midrule
convert rate, (layer=10, $\alpha$=2) & 0.748 & 0.757 & 0.994 & 0.974 & 0.900 & 0.991 & 0.673 & 0.601 & 0.830 & 0.656 & 0.608 & 0.979\\
\midrule
convert rate, (layer=10, $\alpha$=3) & 0.952 & 0.963 & 0.998 & 1.000 & 0.995 & 1.000 & 0.795 & 0.731 & 0.966 & 0.936 & 0.914 & 0.997\\
\bottomrule
\end{tabular}
\vspace{-0.25cm}
\end{table*}


\subsection{Unsupervised Token Translation}
\subsubsection{Method}
We used MDS to force m-BERT to decode in target language $L_2$ different from input language $L_1$. For example, we fed an English sentence into m-BERT, extracted the hidden representation of each token at layer $l$, and then added weighted $D^l_{L_{1}\veryshortarrow L_{2}}$ to the hidden representation of each token. The modified hidden representation $\hat{\bm{h}}^l_{x_i}$can be written as $$\hat{\bm{h}}^l_{x_i} = \bm{h}^l_{x_i} + \alpha D^l_{L_{1}\veryshortarrow L_{2}}$$ where $\alpha$ is a hyperparameter. Finally, we used the modified hidden representation to forward to the remaining layers after $l+1$-th layer and predicted tokens by computing masked language model probabilities, although we did not mask any token at input sequence. 
Surprisingly, we can get predicted tokens in another language in this way, and most of them are the token level translation of the input English words.
Interestingly, we observe that as $\alpha$ grows from 0, the model decodes more tokens to target language and never decodes the tokens which not belong to $L_1$ and $L_2$. When given a negative $\alpha$, the model always decodes the tokens belong to $L_1$. 
The sample outputs are reported in Table~\ref{tab:ex}. 

\subsubsection{Evaluation}
Although we could force the model to decode in target language, the quality of translation was not acceptable. Instead of measuring the translation quality, we calculateed how many tokens can be converted from source language to target language.
We used BLEU-1 and convert rate as our evaluation metrics. Convert rate (CR) is defined as following
$$ \mathrm{convert \ rate} = \frac{\text{\# of } y \in V_t-V_s}{\text{\# of } y - \text{\# of } y \in V_s\cap V_t}$$
where $y$ is the output tokens of the model, $V_s, V_t$ is the token set of the source and target language. 
Shared tokens in both vocabularies were not taken into account therefore excluded from the numerator and denominator term.

We computed the BLEU-1 score using ground truth reference translation of and the convert rate each sentence in XNLI test-set. We showed the result of the best setting (layer=11, $\alpha$=3), and the best score of each language pair and its weight and $\alpha$ in Table~\ref{tab:mt}.

Although the translation results were not comparable to any unsupervised translation methods that usually apply denoising pretraining before translation~\citep{kim2018improving}, it shows strong evidence that we can manipulate the language-specific information in the representations by MDS, and force m-BERT to switch to another language. The covert rate shows that most of the tokens can be converted into another language. The BLEU-1 scores also show that many of the converted tokens are their translated words without considering the fluency of the converted sequence.

\begin{figure}[ht]
    \centering
    \includegraphics[width=1.0\linewidth]{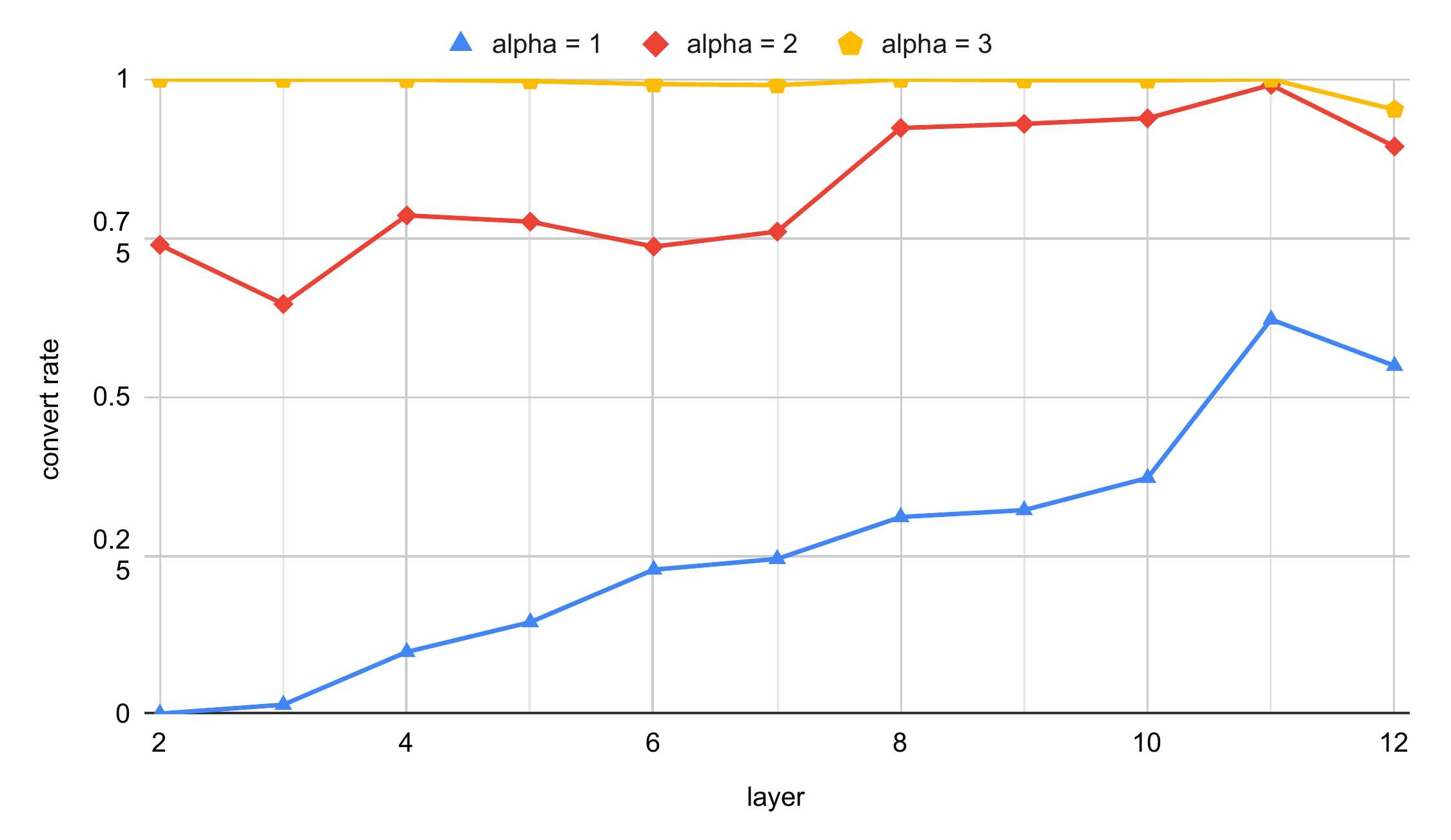}
    \caption{Convert rate on \textit{en $\veryshortarrow$ el} data when applying different $\alpha$ on different layers for MDS.}
    \label{fig:convert}
\end{figure}

\begin{figure}[ht]
    \vspace{-0.25cm}
    \begin{subfigure}[b]{\linewidth}
        \includegraphics[width=1.0\linewidth]{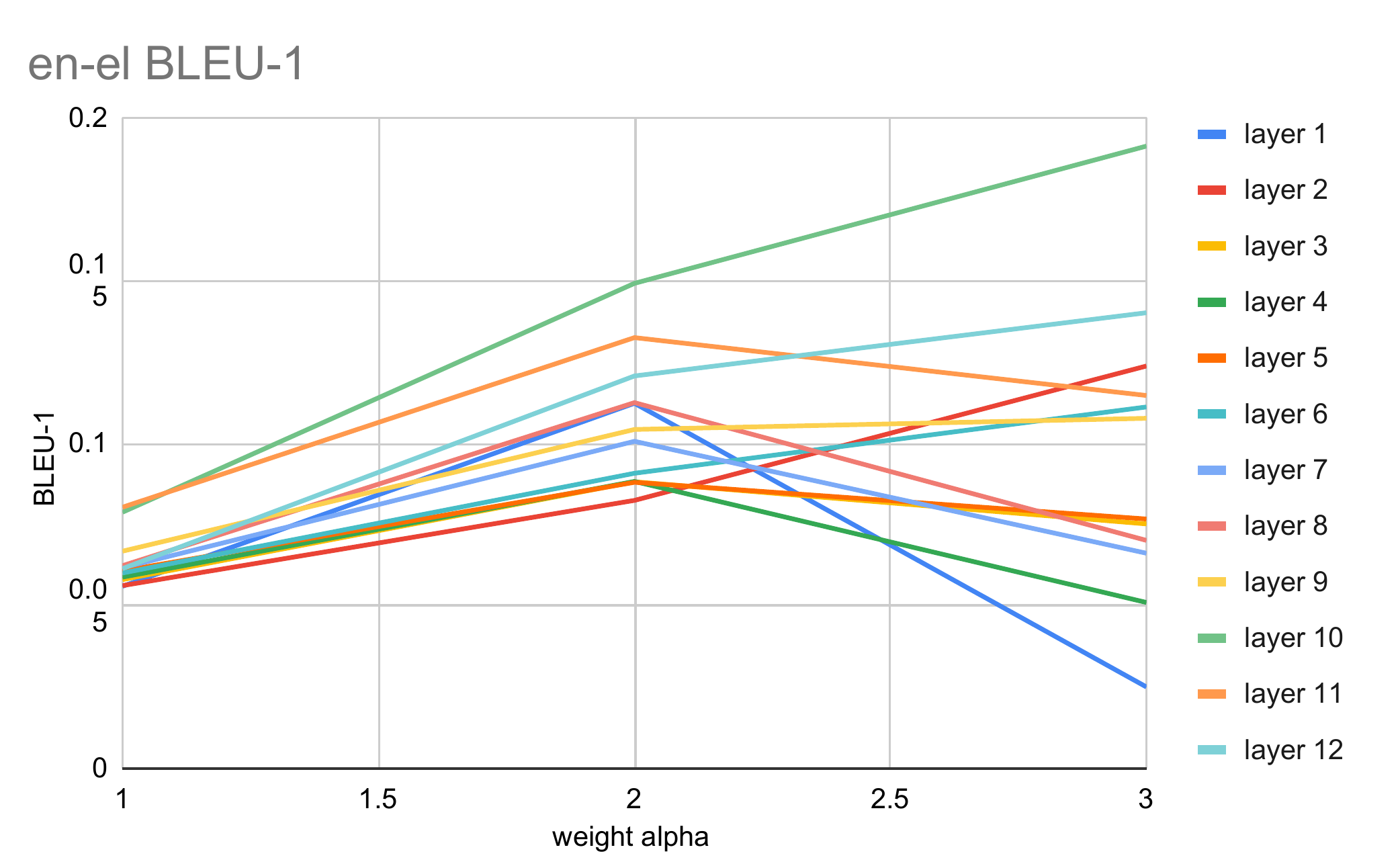}
        \subcaption{x-axis presents different weight $\alpha$.}
    \end{subfigure}
    \begin{subfigure}[b]{\linewidth}
        \includegraphics[width=1.0\linewidth]{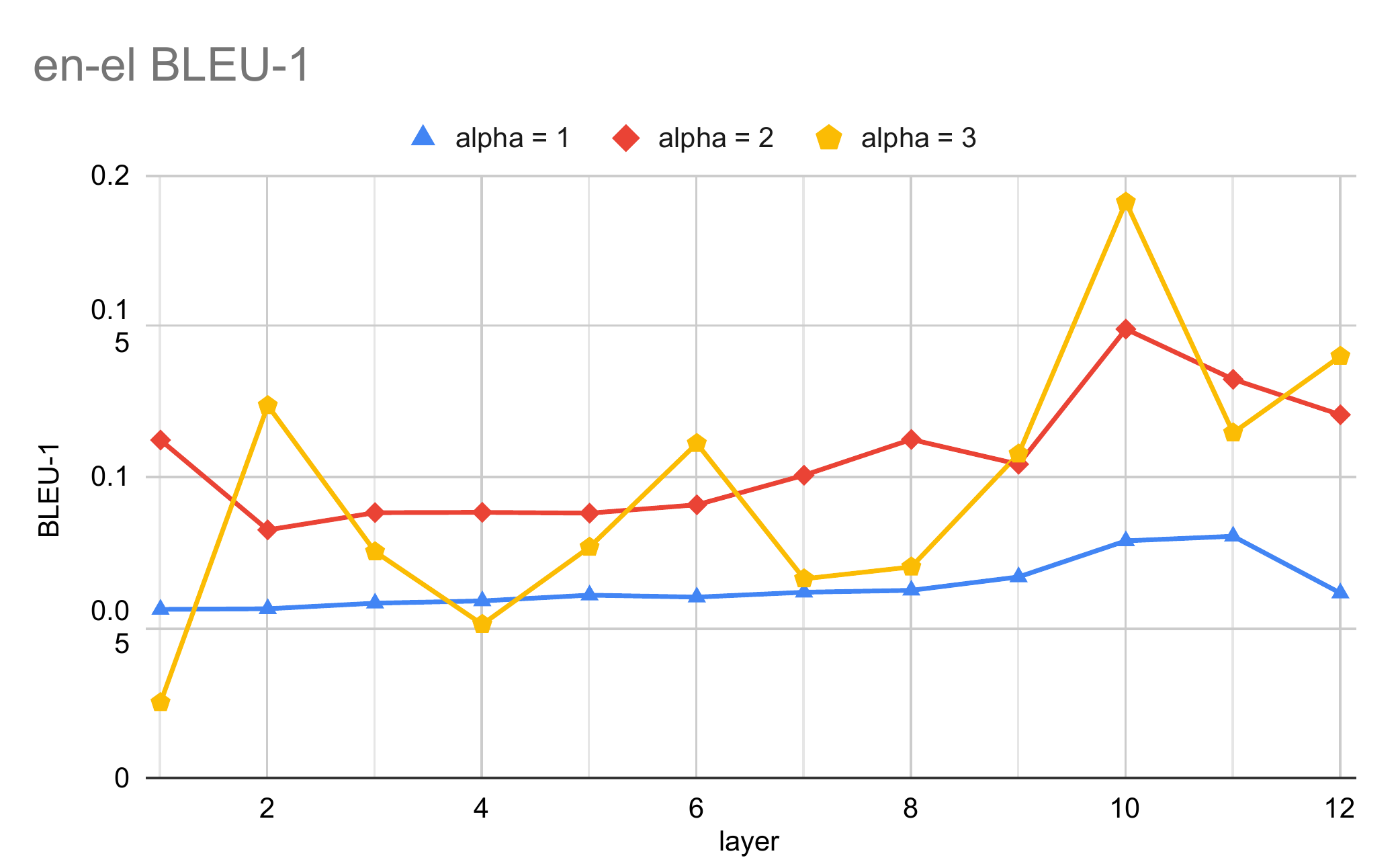}
        \subcaption{x-axis presents different layers.}
    \end{subfigure}
    \caption{The direction of change on BLEU-1 of en $\veryshortarrow$ el unsupervised token translation when tuning the shifting weight $\alpha$ and selecting different layers.}
    \label{fig:bleuchange}
    \vspace{-0.25cm}
\end{figure}

We present an example of \textit{en $\veryshortarrow$ el} in an Figure~\ref{fig:convert} and ~\ref{fig:bleuchange} to show that how convert rate and BLEU-1 score changed with different weights $\alpha$ and different layers. 
We observed the influences of weight increase on convert rate were monotonic increase, there are more converted tokens as weight increasing.
Although, the influences of weight increase on BLEU-1 score were mixed, it is worth mentioning that in the last few layers (10 or 11), the BLEU-1 of most languages obviously rose when $\alpha$ was set to $3.0$ (also shown in the best layer row in Table~\ref{tab:mt}). 
It indicates that the last few layers may be better for disentangling language-specific representations, which is consistent with the observation in the literature that the last few layers contain more language-specific information for predicting masked words~\citep{pires:19}.

\begin{table}[t]
    \label{tab:common}
    \caption{Size of token set and size of English token set intersection with another language token set.}
    \centering
    \footnotesize
    \setlength\tabcolsep{3pt}
    \begin{tabular}{c|ccccccc}
    \toprule
	& en & de & fr & el & zh & ur & sw \\ \midrule
	$|V_{\text{lang}}|$ & 9140 & 9212 & 8552 & 3189 & 3866 & 4085 & 5609 \\ \midrule
    $|V_{en}\cap V_{\text{lang}}|$ & 9140 & 3230 & 3911 & 1696 & 1325 & 1549 & 2970 \\
	\bottomrule
    \end{tabular}
    \vspace{-0.5cm}
\end{table}

\begin{table*}[ht]
\centering
\caption{Experiment results of applying mean difference shift to m-BERT on XNLI test-set. Shifting weights are selected according to the best results on XNLI dev-set.}
\label{tab:xnli}
\footnotesize
\setlength\tabcolsep{3pt}
\begin{tabular}{c|cccccccccccc|c}
\toprule
& en & de & es & ar & el & fr & hi & ru & th & tr & vi & zh & avg w/o en \\
\midrule
finetune all layers & 80.86 & 67.05 & 70.6 & 59.94 & 56.67 & 70.42 & 46.65 & 66.97 & 41.70 & 50.10 & 40.68 & 67.20 & 58.00 \\
\midrule
finetune last 6 layers & 82.16 & 69.34 & 74.73 & 63.91 & 61.52 & 72.95 & 51.00 & 69.46 & 48.94 & 57.23 & 40.76 & 69.96 & 61.80 \\
+ MDS & - & \textbf{69.72} & 74.57 & \textbf{64.53} & \textbf{62.02} & 72.42 & \textbf{51.28} & \textbf{69.56} & \textbf{49.90} & \textbf{57.45} & \textbf{43.07} & \textbf{70.16} & \textbf{62.24} \\
\midrule
shifting weight &   & 1.5 & 0.2 & 1.0 & 0.9 & 0.7 & 0.3 & 1.0 & 0.4 & 0.4 & 0.5 & 2.0 \\
\bottomrule
\end{tabular}
\vspace{-0.25cm}
\end{table*}

\subsection{Cross-lingual Transfer}
In previous works~\citep{pires2019multilingual, conneau2019cross}, people attempted to solve cross-lingual transfer tasks like XNLI in a purely zero-shot setting by training on English dataset and then testing on the dataset of other languages. 
However, as we have shown, the representation has language-specific information, so the training language and the testing languages still have mismatched representations. 
We tried to eliminate the difference in language-specific information using MDS. 
For XNLI task, we finetuned the m-BERT model on the English dataset while fixing the first six layers of it, as the transfer results of finetuning the last 6 layers are slightly better than finetuning the whole BERT model. 
On the testing set, we fed the tuned model with datasets from a new language and added MDS to the hidden representations of the 6-th layer of our model. 
And then we can get prediction by forwarding the modified representation to the remaining layers after the 6-th layer. 

When applying MDS, we used different shifting weights $\alpha$ for different languages considering the sensitivity may differ among languages. We found the best shifting weight $\alpha$ on XNLI development sets and reported the accuracy when applying the same weight on XNLI testing sets in Table~\ref{tab:xnli}. We got a slight improvement on the testing sets from most of the languages, except Spanish (es) and French (fr). The results show that by simply using MDS on the XNLI dataset, the language-specific information can be eliminated to some extent, and the modified representations can preserve more semantic information instead of mixed with language-specific information. 

As for Spanish (es) and French (fr), MDS did not bring any advantages. 
We speculated that the results came from the higher subword overlapping with English. 
Some shared subwords have the same meaning no matter in English or target language, but others don't.
Because MDS used one mean vector to present language-specific representation without considering the difference among vocabularies, simply apply MDS to all vocabularies of these languages may hurt performance instead. A more delicate mechanism should be considered to apply different shifts for different vocabularies. 
We leave this research topic for future work.

\section{Conclusion}
In this paper, we compare non-contextualized word embeddings with m-BERT. We find out that the cross-lingual ability of m-BERT has been learned from longer dependency (hundreds of tokens) instead of local co-occurrence information, and a massive amount of data is necessary. We also find out that we could make m-BERT align better between languages by just making a shift to its embeddings, which is demonstrated in the context of unsupervised translation and the performance improvement of cross-lingual transfer learning. Our work opens a new direction that strengthens cross-lingual ability different from previous alignment-based methods.     

\bibliographystyle{acl_natbib}
\bibliography{acl2020}

\end{document}